# An Ontology to support automated negotiation

Susel Fernandez
Nagoya Institute of Technology
Gokiso-cho, Showa-ku, Nagoya,
Aichi, Japan, 466-8555
+81-052-735-7968
susel.fernandez@uah.es

Takayuki Ito
Nagoya Institute of Technology
Gokiso-cho, Showa-ku, Nagoya,
Aichi, Japan, 466-8555
+81-052-735-7968
ito.takayuki@nitech.ac.jp

## ABSTRACT
In this work we propose an ontology to support automated negotiation in multiagent systems. The ontology can be connected with some domain-specific ontologies to facilitate the negotiation in different domains, such as Intelligent Transportation Systems (ITS), e-commerce, etc. The specific negotiation rules for each type of negotiation strategy can also be defined as part of the ontology, reducing the amount of knowledge hardcoded in the agents and ensuring the interoperability. The expressiveness of the ontology was proved in a multiagent architecture for the automatic traffic light setting application on ITS.

## Categories and Subject Descriptors
I.2.4 [**Artificial Intelligence**]: Knowledge Representation Formalisms and Methods – *Ontologies*; I.2.11 [**Artificial Intelligence**]: Distributed Artificial Intelligence – *Intelligent agents, Multiagent systems*.

## 1. INTRODUCTION
Negotiation is a type of interaction between two or more parties intended to reach a beneficial outcome over one or more issues where a conflict exists with respect to at least one of these issues [1]. It is aimed to resolve points of difference, to gain advantage for an individual or collective, or to craft outcomes to satisfy various interests. It is often conducted by putting forward a position and making small concessions to achieve an agreement.

In multiagent systems, negotiation is a key form of interaction that enables groups of agents to arrive at a mutual agreement regarding some belief, goal or plan. Thus, understanding the interactions between the parties during the negotiation process is a very important issue in automated negotiation in multiagent systems.

Recently, Semantic Web technologies have been maturing to make negotiation and agent interactions more flexible and automated. Semantic Web technologies provides explicit meaning to the information available on the Web for automated processing and information integration. Due to its high degree of expressiveness, the use of ontologies is crucial to ensure greater interoperability among software agents and different applications in the Semantic Web. Ontologies provide a common vocabulary in a given domain and allow defining, with different levels of formality, the meaning of terms and relations between them [2], as well as rules in formal semantics to facilitate the inference process. The use of ontologies can help negotiators to better understand the negotiation process. The ontology enables that agents participate in negotiations without prior knowledge of the negotiation mechanism. The ability of exchanging knowledge about it reduces the amount of knowledge hardcoded in the agents [3].

Our proposal is a general ontology expressive enough to cover the knowledge and strategies to support the negotiation process in multiagent systems. The ontology can be connected with some domain-specific ontologies to facilitate the negotiation in different domains, such as ITS, e-commerce, etc. The specific negotiation rules for each type of negotiation strategy are also defined as part of the ontology, reducing the amount of knowledge hardcoded in the agents and ensuring the interoperability.

The remainder of this paper is organized as follows. Section 2 is a review of the state of the art in the use of ontologies for negotiation. Section 3 presents our proposed negotiation ontology in detail. In section 4 a negotiation example in the ITS scenario is explained. The experiments are described in Section 5. Finally the conclusions and lines of future work are summarized in section 6.

## 2. ONTOLOGIES FOR NEGOTIATION
Although there are some previous works related to the development of ontologies for negotiation, most of them are not expressive enough. This is in part due to the fact that in most of the cases the negotiations are not open, that is, agents do not share their negotiation strategies or preferences. Therefore the ontologies can't yet cover all the stages of the negotiation process, being necessary the implementation of decision making algorithms in the agents.

In [3] a general architecture for negotiation process which uses an ontology-based knowledge management system is proposed. The architecture consists of three layers: the negotiation layer that describes the negotiation process between the initiator static agents and the participant mobile agents by using suitable ontologies; the semantic layer, that contains the semantic translator for the case of misunderstanding of the sent messages between the agents; and the last one is the knowledge management systems layer, which is based on the intelligent knowledge base to give the flexibility to their negotiation ontology.

Tamma and Wooldridge proposed a general negotiation ontology [4]. This ontology is intended to capture similarities between the different negotiation mechanisms. This kind of generic description can be used as classification framework that permits the analysis of the negotiation protocols available, and the development of new ones. The negotiation ontology involves general concepts and relations between them in the negotiation scenarios, such as the negotiation process, negotiation protocol, party, object, offer, and negotiation rules. Using this ontology, agents can participate in different negotiation protocols, since a common format for exchanging information is provided.

In [5] they proposed a distributed knowledge-based multi-agent architecture to support automated negotiation within virtual enterprises. They developed a negotiation ontology, based in [4]. It handles different products or services, considering their properties and enables users to have an accurate representation of the surrounding negotiation world. To ensure all of the mentioned points the main concepts of this approach are *User*, *MarketAgent*, *Auctioneer*, *Offer*, *Trade* and *AuctionType*. In this system, a special decision support algorithm is implemented in the agents to recommend the best negotiation strategy.

In [6] an upper level ontology for rational negotiation is proposed. This ontology was designed to be capable to model arbitrary types of negotiation dialogues by providing the generic namespace for that. The models of the specific negotiation types refine the namespace by providing the instances of its concepts, applying additional restrictions on the properties, possibly adding new subclasses and properties.

The work in [7] proposes a framework to support the interoperability in networked enterprise environments. The approach tackles the issue of semantic heterogeneity by introducing ontologies as the main support in the negotiation process. It includes a methodology for the definition of the different processes for capturing knowledge and modelling the environment, a formal negotiation model to represent the negotiation steps, strategies, and an infrastructure for handling, formalizing and persisting negotiation activities.

The work in [8] presents a novel application of Semantic Web technologies for the facilitation of e-Negotiation processes. They discuss how the elicitation of negotiation issues, alternatives, and tradeoff can be streamlined. They proposed a novel methodology for the elicitation of dependencies among negotiation issues from ontologies so that negotiators can focus on tradeoff among interrelated issues, instead of arguing about single issues. A negotiation plan can thus be derived to observe negotiation orders across different issues. As a result, negotiators can have a better cognition of their negotiation tasks and the overall e-Negotiation process can be streamlined.

## 3. THE PROPOSED NEGOTIATION ONTOLOGY

Taking some elements from [6] as basis, we propose a general ontology to support negotiation. The ontology was developed in OWL using Protégé [9] as development tool, with the reasoner Pellet [10]. Fig. 1 shows the main concepts and relations of the negotiation ontology. These elements are explained below.

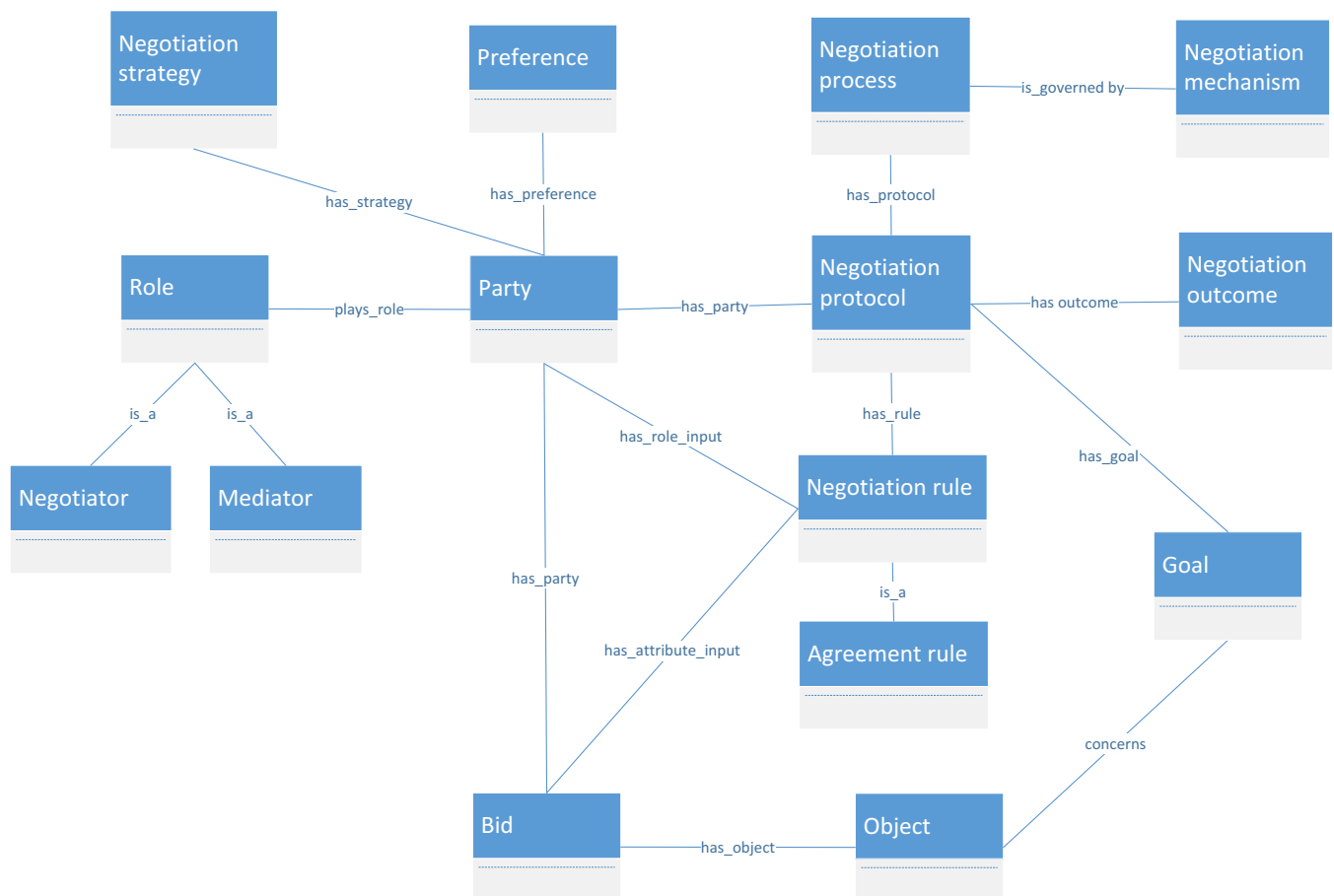

**Figure 1. Fragment of the negotiation ontology**

- *Negotiation Process*: This concept represents the process in which the agents persuade the goal of reaching a common agreement on the negotiation object. The negotiation process is governed by the negotiation mechanism and uses negotiation protocols.
- *Negotiation Mechanism*: This concept represents the components that govern a negotiation process, such as: participants, interaction states, actions, etc.
- *Negotiation Protocol*: This concept aggregates the components which govern the interactions between negotiation parties. The negotiation protocol has an outcome which is associated to a goal accomplishment, concerning a negotiation object.
- *Negotiation Outcome*: This concept represents the result of a negotiation process about a negotiation object.
- *Negotiation Goal*: This concept represents the final objective of the negotiation process, which is concerning a negotiation object.
- *Negotiation Object*: This concept represents the item or set of items about which the different parties are negotiating.
- *Negotiation Rule*: This concept represents the negotiation rules that guide a negotiation protocol. Through this concept, the parties and attributes of a negotiation algorithm can be defined.
- *Agreement Rule*: This concept represents a type of negotiation rule to specify the parameters that are necessary to reach an agreement in a negotiation process.
- *Negotiation Party*: This concept represents the participants in the negotiation process. A negotiation party plays a specific role in the negotiation process.
- *Preference*: This concept represents the preferences that the different parties want to publish during the negotiation process. Sharing some of their preferences can help the agents to reach agreements easily in some negotiation processes.
- *Negotiation Strategy*: This concept represents the internal behavior of a negotiation party that plays a role in a negotiation process. A party employs the strategy to reason about the next movement in the negotiation process.
- *Role*: This concept represents the role played by a negotiation party. Negotiation roles frame out the goals and the strategies of the parties. In this work we have identified two roles for the negotiation parties: the negotiator and the mediator.
- *Bid*: This concept represents the bids made by the different parties for negotiate about some objects in a negotiation process.

## 4. EXAMPLE OF NEGOTIATION IN ROAD TRAFFIC SCENARIO

Consider a scenario composed of various road intersections, with different traffic lights each. Taking into account the possible routes from origin to destination, cars negotiate the status of the traffic lights to minimize the waiting time at intersections. There is a mediator agent, which is responsible for selecting the best configuration of traffic lights (contract) that benefits all and facilitates the traffic flow. Based on the representation proposed in [11], in this case we would have *m* agents who want to negotiate *n* issues (traffic lights states), for each intersection different vectors are designed $S_i = (c_{i1}, c_{i2},.., c_{in})$, where $S_i$ is a configuration of traffic lights (a bid) and $c_{ij}$ is a constraint that indicates the state of each light (the value of the issue *j* in the bid *i*).

If we start from the knowledge represented in the traffic ontology that we proposed in [12], we can get all the knowledge of each intersection (location, type and status of different traffic signs; characteristics of each road, such as number of lanes, the direction of lanes, information about traffic regulations, etc) querying the ontology. With this information we can infer, for example if at some point we can turn right on an intersection, or if we must give way to a priority vehicle. The knowledge about negotiation is modelled using the ontology showed in Fig.1. In our proposal, the negotiation concepts and traffic concepts (taken from the negotiation and traffic ontologies respectively) are connected via the concept *Object* (which is the abstract concept for the description of the issues to negotiate). In this case, the object of negotiation will be the traffic light states.

In this scenario we can make different queries to the ontology, to retrieve some information such as for example:

- The location of the traffic lights along the route (IsOn(?traffic_light, ?road)).
- The location of the next traffic light along the route (TrafficLightAtPoint(NextRoutePoint(?route_point, route?)).
- Know if a traffic light may have the right_green state (has_Part(?light, ?right_green)).
- Know the relation between road segments location (IsAtRight(road_seg1, road_seg2)) or (IsAtLeft(road_seg1, road_seg2)).
- Know if a road segment is connected to an intersection (ConnectedTo(road_seg, intersection)).

In this work, there is a mediator agent on each intersection to control the negotiation process. The mediator agent prepares different possible configurations of traffic lights as bids, and then, the vehicle agents vote for their preferred configuration. In the proximity of the intersection, each vehicle retrieve the available bids about the next traffic light configuration from the ontology and decide which is the most convenient for their objectives. At the end, the most voted configuration will win. With this strategy we guarantee to give the highest priority to the most congested roads.

An example of one route along four intersections, with 4 traffic lights each is shown in Fig. 2. In this example, let's suppose that the car 1 wants to go from point A to point B following the route marked with red dashed lines. The timeline at the bottom of the figure represents the sequence of moments in which the car1 reach the different intersections and vote for the traffic light configuration.

Table 1 shows two possible configurations offered by mediator agents about the traffic lights in the four intersections of the example. As we can see in Fig. 2, at the time T1 (intersection 1), there are five cars reaching the intersection. The different car agents vote for their favorite configuration and, in this case three agents voted for the first configuration and the other two agents voted for the second one. Thus the winner configuration was that in which TL1 and TL3 are green while TL2 and TL4 are red. At this point, the car 1 is able to turn right following the route without waiting.

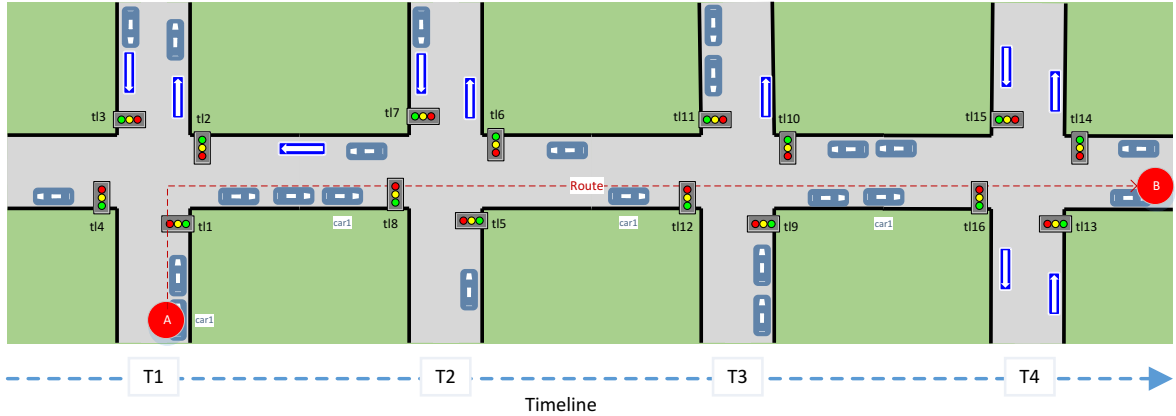

**Figure 2. Example of traffic negotiation scenario**

**Table 1. Example of possible traffic light configurations**

| Intersection | Bid 1 | Bid 2 |
|---|---|---|
| Int1 | Configuration1=<br>(TL1.green,<br>TL2.red,<br>TL3.green,<br>TL4.red) | Configuration2=<br>(TL1.red,<br>TL2.green,<br>TL3.red,<br>TL4.green) |
| Int2 | Configuration1=<br>(TL5.green,<br>TL6.red,<br>TL7.green,<br>TL8.red) | Configuration2=<br>(TL5.red,<br>TL6.green,<br>TL7.red,<br>TL8.green) |
| Int3 | Configuration1=<br>(TL9.green,<br>TL10.red,<br>TL11.green,<br>TL12.red) | Configuration1=<br>(TL9.red,<br>TL10.green,<br>TL11.red,<br>TL12.green) |
| Int4 | Configuration1=<br>(TL13.green,<br>TL14.red,<br>TL15.green,<br>TL16.red) | Configuration1=<br>(TL13.red,<br>TL14.green,<br>TL15.red,<br>TL16.green) |

At the time T2, 6 vehicles reach the intersection. Four of them voted for the second configuration and the other two voted for the first one. Thus the winner configuration was that in which TL6 and TL8 are green while TL5 and TL7 are red, and the car 1 can keep moving to the third intersection without waiting.

At the time T3, 7 vehicles reach the intersection. In this case, configuration 2 received 3 votes while 4 cars voted for configuration 1 (TL9 and TL11 in green while TL10 and TL12 in red). That means that the other road at crossing is more congested, and car 1 has to wait. Finally at the time T4, the three vehicles that reach the intersection voted for the second configuration and that configuration wins. That means that car 1 can keep moving until the end of the route.

The agents make queries to the ontology about the road status, the possible states of traffic lights, and the possible configurations of traffic lights (bids) to vote for. Each bid will have a utility value, which is given by the agent preferences.

In the case of car agents, the utility function can take into account different factors, such as minimizing the number of turns in the path or the distance traveled by the car. In the case of the mediator agent on each intersection, the utility of each configuration will be the number of votes that the configuration receives from the car agents. Thus, the most voted configuration will have the highest utility value. To avoid an inappropriate behavior of car agents, they can only vote for one configuration at an intersection, but the utility value of each configuration is public. Thus, knowing the utility, they can change their votes at any time before the application of the configuration.

We can also use the ontology to model the semantics of a negotiation protocol. The following is an example of a negotiation protocol called *TrafficLightSign* with two constraints in this example, one for *has_Actor* relation, and the other for the *has_Object* relation:

```
<owl :Class rdf:about="&TrafficLightSign">
 <rdfs: subClassOff rdf: resource="&Protocol"/>
  <rdfs: subClassOff>
  <owl :Restriction>
    <owl :onProperty rdf:resource="&hasActor">
    <owl:minCardinality
      rdf:datatype="nonNegativeInteger">2
    </owl:minCardinality>
    <rdfs:comment>
      This negotiation needs at least 2
      participants
    <rdfs:comment>
  </owl :Restriction >
  <owl :Restriction >
    <owl :onProperty rdf:resource="&hasObject">
    <owl:hasValue
      rdf:resource="&TrafficOnto;TrafficLight"/>
    <rdfs:comment>
       The negotiation object is a traffic light
       state
    <rdfs:comment>
   </owl :Restriction >
 </rdfs: cubClassOff>
</owl :Class>
```

The objects to negotiate are defined as the traffic lights on intersections along the route of the car. The following is an example of one object definition to negotiate (*TrafficLight*). The example shows a definition of a multiple elements of this type to be negotiated. Each of them has four attributes. The attributes represents the four possible states of the traffic light:

```
<owl :Class rdf:ID="&TrafficLight">
  <rdfs: subClassOff rdf: resource="&Object"/>

  <rdfs: subClassOff>

  <owl: Restriction>
    <owl: onProperty
       rdf: resource="&NumberOfItems">
    <owl: hasValue rdf:resource="&Multiple">
  </owl: Restriction >

  <owl: Restriction>
    <owl: onProperty
       rdf: resource="&NumberOfAttributes">
    <owl: hasValue rdf:resource="4">
  </owl: Restriction >
 </rdfs: cubClassOff>
</owl: Class>
```

For each possible configuration of the lights on the intersection, the mediator agent will create a bid for negotiation. The agents also keep a list of possible configurations associated to different routes ordered by utility value. Knowing the configuration offers for the next intersection and the state of the voting for each configuration, the car agent check its preferences and decide the configuration to vote for. That configuration should be the most convenient for the car to accomplish the route in a shorter period of time.

As we mentioned before, for the mediator agent, the utility of the bid will be considered as the number of vehicles that voted for that configuration. A small fragment of the definition of a possible bid from the mediator_agent1 with utility value of 9 is presented below:

```
<owl: Class rdf:ID="&MediatorAgent1_Bid1">
 <rdfs: subClassOff rdf: resource="&Bid"/>
  <rdfs: subClassOff>
   <owl: Restriction>
    <owl: onProperty rdf:resource="&hasParty">
    <owl: hasValue
       rdf: resource="&MediatorAgent1">
   </owl: Restriction>

   <owl: Restriction>
    <owl: onProperty rdf:resource="&hasUtility">
    <owl: hasValue rdf:resource="9">
   </owl: Restriction>

   <owl: Restriction>
    <owl: onProperty rdf:resource="&hasObject">
    <owl: hasValue rdf:resource="&TrafficLight1">
   </owl: Restriction>

   <owl: Restriction>
    <owl: onProperty rdf:resource="&hasObject">
    <owl: hasValue rdf:resource="&TrafficLight2">
   </owl: Restriction >
     .
   </rdfs: cubClassOff>
</owl: Class>
```

Having made the count of the voting, the mediator agent decides the winner bid. Through negotiation rules we can define different elements of the negotiation step as input and output parameters of different algorithms used. An example of agreement rule definition could be the following:

```
<owl: Class
  rdf:ID="&TrafficLightSignAgreementRule">

<rdfs: subClassOff rdf:
     resource="&AgreementRule"/>
</owl: Class>

<owl: ObjectProperty rdf:ID="&hasRoleInput">
    <rdfs: Domain rdfs: resource
       ="&TrafficLightSignAgreementRule">
    <rdfs: Range rdfs: resource ="&Party">
</owl: ObjectProperty>

<owl: ObjectProperty rdf:ID="&hasAtributeInput">
    <rdfs: Domain rdfs: resource
       ="&TrafficLightSignAgreementRule">
    <rdfs: Range rdfs: resource ="&Bid">
</owl: ObjectProperty>

<owl: ObjectProperty rdf:ID="&hasAtributeOutput">
    <rdfs: Domain rdfs: resource
       ="&TrafficLightSignAgreementRule">
    <rdfs: Range rdfs: resource ="&Bid">
</owl: ObjectProperty>
```

## 5. Experiments

A set experiments were designed to validate the expressiveness of the proposed negotiation ontology in the specific problem of the automatic traffic light settings on intelligent transportation systems.

A multiagent architecture for simulation was deployed using the Java Agent Development Framework (JADE) [13], and as query language to retrieve information from the ontology, SPARQL [14] was used.

In this work, a total of 150 experiments were conducted, with 300 vehicles for which different traffic scenarios were defined. These scenarios were composed of 50 routes within a radius of 10 km, with 10 intersections regulated by traffic lights (40 traffic lights).

On each simulation, the time spent for the vehicles to reach their destination, following the specific route was computed. This calculation was made taking into account the length of the roads and the speed of the vehicles, which was defined at the beginning and it was assumed to be constant during the whole path.

The chosen initial configuration for traffic lights, and the position and route of the vehicles was selected randomly. The duration of each traffic light was adjusted taking into account the level of congestion of the roads involved on each intersection.

For testing the effectiveness of the negotiation using the ontology in these scenarios, the experiments were performed in two phases: the first one is a simulation of the system applying the negotiation for the traffic light settings; and the second one is without the negotiation. The comparison of the results of both phases of the experiments is shown in Table 2.

**Table 2. Overall results of the experiments**

| Parameter | Value |
|---|---|
| Number of experiments | 150 |
| Number of vehicles | 300 |
| Number of routes | 50 |
| % of vehicles that gained time | 80% |

| Average of time gained | 152 seconds |
|---|---|
| % of vehicles that not gained time | 14% |
| % of vehicles that lost time | 6% |

The table shows the % of vehicles that gained time using the negotiation ontology for the automatic synchronization of traffic lights; the average of time gained by vehicle; the % of vehicles that did not gain time and the % of vehicle that lost time. As we can see in the table, 80 % of vehicles experienced a gain of time, and the average of gained time was approximately 152 sec. The vehicles that did not gain time in the experiments (19%) were those that traveled through routes with some parts in congested roads and some parts in non-congested roads. The % of vehicles that experienced a loss of time was only 6%. The vehicles that lost time were those that traveled along non-congested routes.

This loss of time occurred because in the experiments the route was fixed before starting and it was not changed. However in practice, with this system, the vehicles can dynamically change their route, taking into account the information about the possible configurations of the traffic lights and the congestion level of the intersections along the different routes. This would allow them to avoid routes in which they could lose time.

## 6. Conclusions

This paper presents a general ontology to support negotiation in multiagent systems. The global objective of the work is the development of an ontology expressive enough to cover the knowledge and strategies of negotiation to support the negotiation process in multiagent systems. This ontology can be connected with some domain-specific ontologies to facilitate the negotiation in different domains, such as ITS, e-commerce, etc. The specific negotiation rules for each type of negotiation strategies can also be defined as part of the ontology, reducing the amount of knowledge hardcoded in the agents and ensuring the interoperability.

The ontology has been tested in several experiments in the scenario of automatic synchronization of traffic lights in a multi-agent architecture for ITS. In this scenario a very simple negotiation mechanism has been applied which consists of a vote of the preferred configuration by the vehicles approaching an intersection, from a series of possible configurations offered by a mediator agent at the intersection. At the end of the negotiation, the most voted configuration wins, giving the highest priority to the traffic on the most congested routes.

The experiments show that the negotiation ontology helps to improve the process of traffic light synchronization, providing information about the current traffic situation and the rules to guide the negotiation process. In general, having an ontology allows the capture and analysis of knowledge concerning different specific domains as well as the different algorithms and rules of negotiation. The knowledge is managed in a structured and readable form, which facilitates interoperability between different trading venues in multiple domains. On the other hand, it facilitates the explanation of the negotiation process too.

However, so far the ontology has only been tested in this scenario, with a very simple negotiation mechanism. Therefore, as future work we intend to incorporate more negotiation mechanisms with different strategies and their agreement rules to the ontology, and also test them in more complex scenarios.

## 7. ACKNOWLEDGMENTS

This work was partially supported by Research and Development on Utilization and Fundamental Technologies for Social Big Data by NICT (National Institute of Information and Communications Technology), the Fund for Strengthening and Facilitating the National University Reformations by Ministry of Education, Culture, Sports, Science, and Technology, Japan, and by the Spanish Ministry of Economy, Industry and Competitiveness grant TIN2016-80622-P (AEI/FEDER, UE).

## 8. REFERENCES

[1] Fisher, Roger; Ury, William (1984). Patton, Bruce, ed. Getting to yes : negotiating agreement without giving in (Reprint ed.). New York: Penguin Books.

[2] Studer, R; Benjamins, R.; Fensel, D. Knowledge Engineering: Principles and Methods. In: Data and Knowledge Engineering, 1998, v.25, n.1-2, pp.161-197.

[3] Saad, S; Zgaya, H and Hammadi, S (2008c). The Flexible Negotiation Ontology-based Knowledge Management System: The Transport Ontology Case Study, In proceedings of the IEEE Int. Conf. on Information & Communication Technologies: from Theory to Applications (ICTTA), 2008.

[4] Tamma, V., Phelps, S., Dickinson, I., & Wooldridge, M. (2005). Ontologies for supporting negotiation in e-commerce. Engineering applications of artificial intelligence, 18(2), 223-236.

[5] Koppensteiner, G., Merdan, M., Lepuschitz, W., Reinprecht, C., Riemer, R., & Strobl, S. (2009, December). A Decision Support Algorithm for Ontology-Based Negotiation Agents within Virtual Enterprises. In Future Information Technology and Management Engineering, 2009. FITME'09. Second International Conference on (pp. 546-551). IEEE.

[6] Ermolayev, V., & Keberle, N. (2006, May). A Generic Ontology of Rational Negotiation. In ISTA (pp. 51-65).

[7] Jardim-Goncalves, R., Coutinho, C., Cretan, A., da Silva, C. F., & Ghodous, P. (2014). Collaborative negotiation for ontology-driven enterprise businesses. Computers in Industry, 65(9), 1232-1241.

[8] Chiu, D. K., Cheung, S. C., Hung, P. C., & Leung, H. F. (2005, January). Facilitating e-negotiation processes with semantic web technologies. In Proceedings of the 38th Annual Hawaii Int. Conf. on System Sciences. IEEE.

[9] Protégé. Available online: http://protege.stanford.edu/ (Access on December 10th , 2016)

[10] Pellet. Available online: http://clarkparsia.com/pellet/ (Access on July 15th , 2015)

[11] Takayuki Ito, Mark Klein, and Hiromitsu Hattori. Multi-issue negotiation protocol for agents: Exploring nonlinear utility spaces. In Proceedings of the 20th Int. Joint Conf. on Artificial Intelligence (IJCAI07), pp. 1347–1352, 2007.

[12] Fernandez, S.; Hadfi, R.; Ito, T.; Marsa-Maestre, I.; Velasco, J.R. Ontology-Based Architecture for Intelligent Transportation Systems Using a Traffic Sensor Network. Sensors 2016, 16, 1287.

[13] JADE. Available online: http://jade.tilab.com (Access on December 20th , 2016)

[14] SPARQL. Available online: http://sparql.org/ (Access on January 9th , 2017).